# Multi-Representational Learning for Offline Signature Verification using Multi-Loss Snapshot Ensemble of CNNs

Saeed Masoudnia[a], Omid Mersa[a], Babak Nadjar Araabi[a], Abdol-Hossein Vahabie[b], Mohammad Amin Sadeghi[a] , and Majid Nili Ahmadabadi[c]

[a] Machine Learning and Computational Modelling Lab, Control and Intelligent Processing Center of Excellence, School of Electrical and Computer Engineering, College of Engineering, University of Tehran, Tehran, Iran.

[b] School of Cognitive Sciences, Institute for Research in Fundamental Sciences (IPM), Tehran, Iran.

[c] Cognitive System Lab, Control and Intelligent Processing Center of Excellence, School of Electrical and Computer Engineering, College of Engineering, University of Tehran, Tehran, Iran.

## Abstract

Offline Signature Verification (OSV) is a challenging pattern recognition task, especially in the presence of skilled forgeries that are not available during training. This study aims to tackle its challenges and meet the substantial need for generalization for OSV by examining different loss functions for Convolutional Neural Network (CNN). We adopt our new approach to OSV by asking two questions: 1. which classification loss provides more generalization for feature learning in OSV? , and 2. How integration of different losses into a unified multi-loss function lead to an improved learning framework?

These questions are studied based on analysis of three loss functions, including cross entropy, Cauchy-Schwarz divergence, and hinge loss. According to complementary features of these losses, we combine them into a dynamic multi-loss function and propose a novel ensemble framework for simultaneous use of them in CNN. Our proposed Multi-Loss Snapshot Ensemble (MLSE) consists of several sequential trials. In each trial, a dominant loss function is selected from the multi-loss set, and the remaining losses act as a regularizer. Different trials learn diverse representations for each input based on signature identification task. This multi-representation set is then employed for the verification task. An ensemble of SVMs is trained on these representations, and their decisions are finally combined according to the selection of most generalizable SVM for each user.

We conducted two sets of experiments based on two different protocols of OSV, i.e., writer-dependent and writer-independent on three signature datasets: GPDS-Synthetic, MCYT, and UT-SIG. Based on the writer-dependent OSV protocol, On UT-SIG, we achieved 6.17% Equal Error Rate (EER) which showed substantial improvement over the best EER in the literature, 9.61%. Our method surpassed state-of-the-arts by 2.5% on GPDS-Synthetic, achieving 6.13%. Our result on MCYT was also comparable to the best previous results. The second set of experiments examined the robustness of our proposed method to the arrival of new users enrolled in the OSV system based on the writer-independent protocol. The results also confirmed that our proposed system efficiently performed the verification of new users enrolled in the OSV system.

## 1. Introduction

Despite impressive advances in computer vision and pattern recognition, several classic problems such as Offline Signature Verification (OSV) remains challenging (Hafemann, Sabourin, & Oliveira, 2017b; E. Zois, Tsourounis, Theodorakopoulos, Kesidis, & Economou, 2019). The OSV system aims to distinguish whether a given signature image is produced by the claimed author (genuine) or by an impostor (forgery). There are three types of forgeries, including random, simple and skilled forgeries. Random forgeries refer to the cases when forger signs without information about the user. In the case of simple forgeries, forger knows the user's name while the user's signature is not available. In the worst case, forger has access to both the user's name and his signature and carefully attempts to forge the signature. In this case, skilled forgeries show high similarity with the user's signature, and thus are harder to verify. The problem of OSV becomes challenging in the presence of skilled forgeries. This challenge is further aggravated in the real scenario of

OSV when a few genuine signatures and no skilled forgeries are available for training (Hafemann et al., 2017b).

Recently, however, automatic feature learning by deep Convolutional Neural Networks (CNNs) has shown its potential to tackle the challenges of OSV (Dey et al., 2017; Hafemann, Sabourin, & Oliveira, 2017a). Due to the small training size, the power of Deep CNNs may not be effectively leveraged in the problem of OSV unless seriously considering regularization. Choosing a suitable loss function is one of most effective learning parameters in the generalization of CNNs (Janocha & Czarnecki, 2017). Substantial need for the generalization in OSV lead researchers to examine different loss functions for CNNs to find which one promotes more generalization.

Two main categories of loss functions were studied for OSV, including metric learning losses and classification (identification) losses. In the first category, several studies (Berkay Yilmaz & Ozturk, 2018; Dey et al., 2017; Rantzsch, Yang, & Meinel, 2016; Soleimani, Araabi, & Fouladi, 2016; Xing, Yin, Wu, & Liu, 2018) suggested several metric learning losses for feature learning in CNNs and a threshold-based verification was then used to classify whether an input signature is genuine or forgery. While the second category includes the studies (Hafemann, Sabourin, & Oliveira, 2016b; Hafemann et al., 2017b; Souza, Oliveira, & Sabourin, 2018) employing classification loss function for feature learning in CNNs and verification is then performed using SVM as a binary classifier. These two categories competed to achieve better accuracy, while the works in the second category mostly obtained more accuracy compared to the first one.

In addition to OSV, these two categories of loss functions were also studied for different classification tasks, and the similar comparative results were reported (Horiguchi, Ikami, & Aizawa, 2016; Janocha & Czarnecki, 2017). The work (Horiguchi et al., 2016) confirmed this result by performing a fair comparison between Cross-Entropy (CE) loss as a classification loss and several state-of-the-art metric learning losses. Furthermore, another study (Janocha & Czarnecki, 2017) obtained similar results, however, proposed that other classification losses rather than CE also have their merits. Due to the superiority of classification losses, we follow the approach of the second category in this research. In OSV literature, a few studies employed CE loss function for feature learning in CNNs. However, as the best of our knowledge, an interesting question on "*which classification loss leads to more generalization in OSV?*" has not been yet studied in OSV.

From another point of view, different loss functions have complementary advantages and limitations (Janocha & Czarnecki, 2017; C. Xu et al., 2016). Ensemble learning (Masoudnia & Ebrahimpour, 2012) can provide a framework for combining the advantages of different loss functions. Moreover, the ensemble framework brings more diversity and regularization and thus more generalization for OSV(Berkay Yilmaz & Ozturk, 2018; Yılmaz & Yanıkoğlu, 2016). However, why diversity in terms of different loss functions in the ensemble can also improve the OSV?

The OSV can benefit from employing different loss functions by which diverse feature sets are learned from signature images. Thanks to these diverse representations, the discrimination power of the OSV can be improved. This is because features that discriminate between genuine signatures of a user and his/her skilled forgeries are different from the features required to discriminate genuine and skilled forgeries of another user.

The diverse representations learned by different classification losses also provide another advantage for the OSV. The methods in the second category employ a CNN trained with a classification loss on the signature identification task. However, the learned representation is then used for signature verification. In the signature application, the identification is simpler than the verification, while the learned features should be transferred to the harder task. These learned features do not necessarily include fine features required for the verification of genuine vs. skilled forgery signatures. Using diverse multi-representation set learned by different loss functions may address this issue since it extends the feature set for the verification. Moreover, the learned diverse feature sets may capture infrequent features (Xie, Deng, & Xing, 2015).

The need for diversity and regularization in the OSV lead us to ensemble approaches. However, the ensemble of CNNs in which each network is independently trained on a different loss function brings heavy computational burden. We attempt to address this limitation using a multi-loss function which eliminates the need for learning each of loss functions from scratch.

In this study, we aim to fill the gap in the literature by examining different loss functions and proposing a novel framework for using a dynamic multi-loss function in CNNs. We adopt a new approach to the OSV problem by asking two questions: 1. which classification loss function provides more generalization for feature learning in OSV? , and 2. How may an integration of different loss functions into a dynamic multi-loss function lead to an improved learning framework?

These questions are studied in the remaining of this paper as follows: The related works on the literature of OSV are reviewed in the following section. Our approach to the first question is presented in section 3. Section 4 and 5 explains our proposed multi-loss ensemble approach for feature learning and the verification method in OSV, respectively. The experimental results are provided in section 6. We finally conclude the results and propose future research directions in the last section.

## 2. Related Works

The related works to our research in three aspects are reviewed in this section: OSV, ensemble learning and multi-loss function.

### 2.1 Offline Signature Verification

The basic challenge in OSV compared to other physical biometrics such as fingerprint or iris (Jabin & Zareen, 2015) is having a high intra-personal variation. This problem occurs in the presence of skilled forgeries, where an impostor carefully attempts to forge the signature. Since there is a high similarity between the genuine signatures of a user and their skilled forgeries, the verification also faces a low inter-class variability challenge. The other Challenges of OSV lie in a limited number of genuine training samples per user and unavailable skilled forgeries in the training phase (Hafemann et al., 2017b). Figure 1 demonstrates some aspects of these challenges on several Persian signature images from UT-Sig dataset (Soleimani, Fouladi, & Araabi, 2017).

We briefly review here the main concepts of OSV and the recently published works. For a more detailed review of OSV see the review paper (Hafemann et al., 2017b; Impedovo, Pirlo, & Plamondon, 2012). The classifiers for OSV can be grouped into two categories: Writer-Dependent (WD) and Writer-Independent (WI).

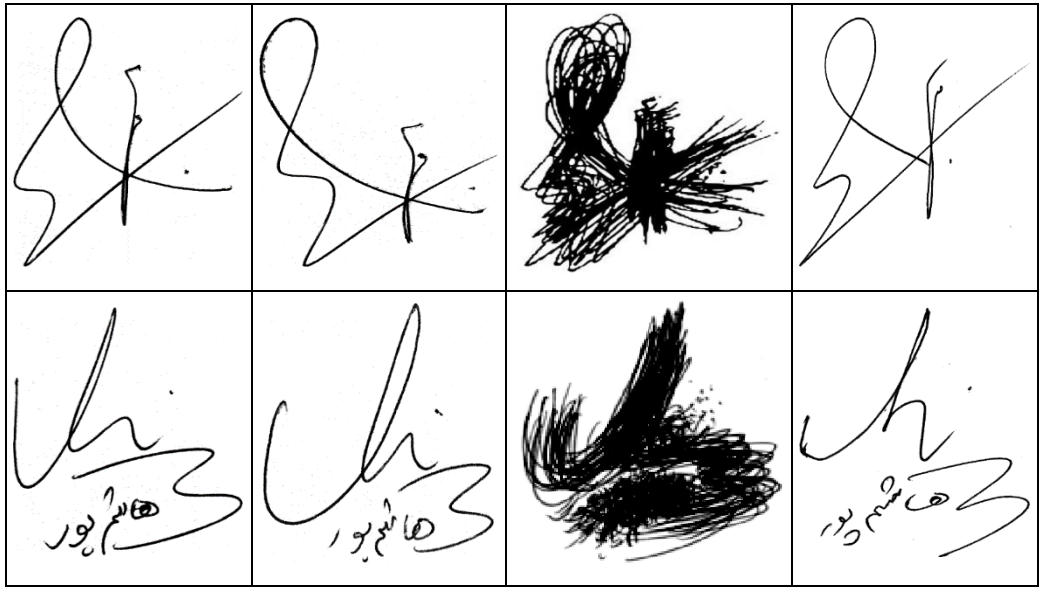

**Fig. 1.** Some signatures of two users in UTSIG (Persian offline signature dataset). Each row shows multiple signatures of the same user. The signature images in the first two columns are two genuine signatures of each user. The third column presents the overlaid genuine signature images of each person, and the fourth column shows a skilled forgery signature for each user. The third column presents a high intra-class variability of the signatures for each user. The high similarity between the genuine signatures in the first two columns and the skilled forgery signatures in the last column also confirms a low inter-class variability.

In the first category, a specialized classifier is trained separately for each user to distinguish the users' signatures from the forgeries. While in the second case only one classifier is trained for all users to determine the authenticity of questioned signatures. WI systems can better handle the problem of small training size compared to WD ones since it uses samples from all authors for only one classifier, but it suffers from missing the writer-specific features.

The most OSV systems (Hafemann et al., 2017b) consists of two main steps: feature extraction and classification. In the feature extraction step, the most researches (Bouamra, Djeddi, Nini, Diaz, & Siddiqi, 2018; Okawa, 2018b; Serdouk, Nemmour, & Chibani, 2016, 2017; Sharif, Khan, Faisal, Yasmin, & Fernandes, 2018; E. N. Zois, Alexandridis, & Economou, 2019; E. N. Zois, Papagiannopoulou, Tsourounis, & Economou, 2018) on OSV have been dedicated to designing hand-crafted feature extraction, while the recent approaches (Berkay Yilmaz & Ozturk, 2018; Hafemann, Oliveira, & Sabourin, 2018; Hafemann, Sabourin, & Oliveira, 2016a; Hafemann et al., 2017a) suggested automatic feature learning by CNNs. In the classification step, two approaches were also used to verify genuine signatures vs. forgeries. These approaches included distance thresholding and SVM classifier. This categorization is summarized in Figure 2.

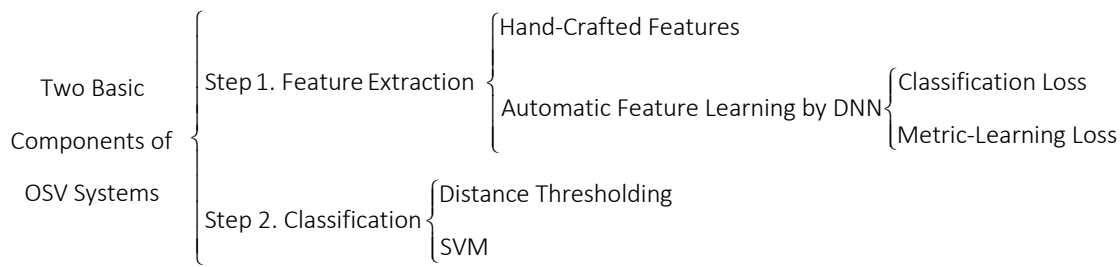

**Fig. 2.** Two essential steps of OSV systems, including feature extraction and classification and also the approaches used for each step.

Based on the type of feature extraction, recent works in two categories of hand-crafted features and feature learning are reviewed in Table 1 and 2 respectively.

Moreover, the automatic feature learning by CNNs are classified in two categories according to the type of loss function: metric learning losses (Berkay Yilmaz & Ozturk, 2018; Soleimani et al., 2016) and identification losses(Hafemann et al., 2018; Hafemann et al., 2016b, 2017a). The outlines of these studies are reviewed in Table 2.

Table 1. The recent studies employed hand-crafted features in the feature extraction step of OSV

| Reference | Features | Classifier | Type |
|---|---|---|---|
| (Vargas, Ferrer, Travieso, & Alonso, 2011) | LBP | SVM | WD |
| (Ooi, Teoh, Pang, & Hiew, 2016) | DRT+PCA | PNN | WD |
| (E. N. Zois, Alewijnse, & Economou, 2016) | Poset-oriented grid features | SVM | WD |
| (Serdouk et al., 2017) | Histogram Of Templates | Artificial Immune System+ SVM | WD |
| (E. N. Zois, Theodorakopoulos, & Economou, 2017) | Archetypal analysis with sparse coding | Thresholding | WD |
| (Alaei, Pal, Pal, & Blumenstein, 2017) | Fuzzy similarity measure and LBP | Thresholding | WD |
| (E. N. Zois, Theodorakopoulos, Tsourounis, & Economou, 2017) | K-SVD dictionary learning and Orthogonal Matching Pursuit | Thresholding | WD |
| (Sharif et al., 2018) | Global + local descriptors in sub-blocks of images + feature selection by GA | SVM | WD |
| (E. N. Zois et al., 2018) | Hierarchical dictionary learning and sparse coding | Thresholding | WD |
| (Okawa, 2018a) | Bag of Visual Words (BoVW) and Vector of Locally Aggregated Descriptors (VLAD) | SVM | WD |
| (Okawa, 2018b) | Fisher vector with fused KAZE features | SVM | WD |
| (Bouamra et al., 2018) | Run-length distributions of binary signature images | One-class SVM | WD |

## 2.2 Ensemble learning and multi-loss function

Regardless of feature extraction approach, a substantial need for diversity and regularization in the OSV has led some researchers to employ ensemble learning (Batista, Granger, & Sabourin, 2012; Bertolini, Oliveira, Justino, & Sabourin, 2010; Yılmaz & Yanıkoğlu, 2016; Zhang, Liu, & Cui, 2016). These studies addressed the challenges of OSV by promoting diversity-induced regularization(Masoudnia, Ebrahimpour, & Arani, 2012b) in ensemble learning. The study in (Zhang et al., 2016) proposed a multi-phase hybrid verification system using both WI-WD classifiers. An unsupervised feature learning was performed by Generative Adversarial Networks (GANs), and several binary classifiers were then trained for each user in the verification phase. The WI discriminator of the GAN and WD classifier were finally fused by Gentle AdaBoost to produce verification decision. Moreover, Yilmaz et al. (Yılmaz & Yanıkoğlu, 2016) proposed an ensemble approach using a score step fusion of classifiers for OSV. The authors described that OSV task benefits from fusing both the global shape and local details.

The ensemble of DNNs, however, has not already been used for OSV since it may not provide an economical and computationally efficient way. Despite this limitation, implicit ensemble view to training process of a

DNN established a new way to promote the generalization without computational burden (Hara, Saitoh, & Shouno, 2016; Huang et al., 2017). Based on this view, Snapshot Ensemble (Huang et al., 2017) proposed employing the stochastic property of Stochastic Gradient Descent (SGD) to make an ensemble based on taking different snapshots from the ResNet CNN along its optimization process with the cyclic learning rates. By reducing the computational burden required for an ensemble of CNNs, this method has been recently used for different applications, however, not already for OSV. A disadvantage of Snapshot Ensemble is using only parametric diversity based on changing learning rates. Relying only on the stochastic property of SGD seems not to be an effective standalone method for generating diversity in an ensemble of DNNs even starting from different initial points (Brown, Wyatt, Harris, & Yao, 2005).

Table 2. The recent papers suggested feature learning by CNNs for OSV.

| Reference | Feature learning | | | | Classifier | |
| --- | --- | --- | --- | --- | --- | --- |
| | Method | Type | Loss Function | | Method | Type |
| | | | Metric Learning | Classification | | |
| (Hafemann et al., 2016b) | The CNN suggested in (Hafemann et al., 2016b) | WI | | ✓ | SVM | WD |
| (Soleimani et al., 2016) | HOG + DMML | WD | ✓ | | Thresholding | WD |
| (Zhang et al., 2016) | Generative Adversarial Network | WI | | ✓ | AdaBoost | WD |
| (Hafemann et al., 2017a) | The CNN (Hafemann et al., 2016b) + 1-bit forgery output | WI | | ✓ | SVM | WD |
| (Souza et al., 2018) | The CNN suggested in (Hafemann et al., 2016b) | WI | | ✓ | Thresholding | WD |
| (Hafemann et al., 2018) | Learning from variable-sized signatures by a CNN through Spatial Pyramid Pooling. | WI | | ✓ | SVM | WD |
| (Berkay Yilmaz & Ozturk, 2018) | Two-channel Siamese CNNs | WI | ✓ | | Thresholding + SVM | WD |
| (Mersa, Etaati, Masoudnia, & Araabi, 2019) | ResNet CNN pretrained on Handwriting classification task | WI | | ✓ | SVM | WD |
| (Younesian, Masoudnia, Hosseini, & Araabi, 2019) | ResNet CNN pretrained on ImageNet | WI | | ✓ | Active learning with SVM | WD |

Instead, switching between different loss functions provides more diversity between the ensemble networks. However, leaning each network by a distinct loss function from scratch requires a heavy computational load. This limitation led researches to study whether integration of different loss functions into a multi-loss function can lead to an improved and cost-effective learning scheme. Several works (McLaughlin, del Rincon, & Miller, 2017; Sun, Chen, Wang, & Tang, 2014) attempted to propose joint classification and metric learning loss functions to combine the strengths of the two losses and address their limitations to improve the discriminative ability of the learned embedding. However, some studies claimed an argument (Chen, Chen, Zhang, & Huang, 2017; Hermans, Beyer, & Leibe, 2017) against the significance of joint classification-verification loss function.

Notwithstanding the argument, The joint loss function could be considered as a special case of multi-loss function. The multi-loss function (C. Xu et al., 2016) was also suggested for CNNs with the aim of addressing regularization. Despite the general title of this method, this framework is not generalizable to simultaneously using any arbitrary loss function. Even though finding the compatible loss functions, a rigorous consideration was taken into account in this work, so that a CNN was pretrained by a single loss function and only fine-tuned then through the multi-loss function. Because of these limitations, this framework could not be generally employed to combine the advantages of different loss functions into a unified system.

### 3. Which losses are better for feature learning in OSV?

Unlike the most researches (Hafemann et al., 2017b) that only used CE, we consider more loss functions, e.g., Hinge loss, and Cauchy-Schwarz Divergence (CSD) loss in this study. We first explain these three losses, discussing their different features and the theoretical hypothesis behind them. These loss functions are further examined independently for training CNNs based on signature identification task, while the learned features are then employed for OSV in our experiments.

- Hinge Loss

  Hinge loss is a margin-based loss which is inspired by the soft-margin SVM, known as L1-SVM. Since L1-SVM is not differentiable, the L2-SVM was proposed (Tang, 2013) which minimizes the squared hinge loss on the samples that violate the margin:

$$\text{Hinge\_loss}(Y,O) = \sum_j \max\left(0, \frac{1}{2} - y_j \cdot o_j\right)^2 \tag{1}$$

where Y and O are the labels and outputs of CNN respectively, and the index $j$ is corresponding to the samples that violate the margin. The advantage of replacing CE with the margin-based loss is the superior regularization effect (Tang, 2013).

- CE loss and CSD loss

CE loss is the most used loss function for classification with DNNs. The motivation behind CE is using entropy between two sampled distributions as an error metric. CE loss works based on KL-divergence of the network's predictions and the target distribution that is calculated as:

$$\text{CE\_loss}(Y,O) = -\sum_j \left[ y_j \log\left(\text{pr}(o_j)\right) \right] \tag{2}$$

where Y and O are the labels and outputs of CNN respectively, and the index j is corresponding to all samples in the training set. The function pr($o_j$) presents the normalized value of *jth* output. Here, we use the soft-max function as:

$$\text{pr}(o_j) = \frac{e^{o_j}}{\sum_i e^{o_i}} \tag{3}$$

An extension for CE loss is recently proposed (Janocha & Czarnecki, 2017), i.e., CSD loss:

$$\text{CSD\_loss}(Y,O) = -\log \frac{\sum_j \text{pr}(o_j) y_j}{\|O\|_2 \|Y\|_2} = \ldots = -\sum_j y_j . \log\left(\text{pr}(o_j)\right) + \log\|O\|_2 \tag{4}$$

As shown in (Janocha & Czarnecki, 2017), CSD loss is approximately equivalent to CE loss regularized with the quadratic entropy of the predictions. CSD loss can also be geometrically interpreted as the cosine distance between two sampled distributions:

$$\text{CSD\_loss}(Y,O) \simeq -\log\left(\cos(Y,O)\right) \tag{5}$$

where cos(Y, O) presents a cosine distance between Y and O. The cosine distance has the advantages of simplicity and bound within the narrow range of [-1,1]. This function is recently used in the loss function of CNNs which leads to a fast and more stable gradient-based optimization algorithm (Wang, Zhou, Wen, Liu, & Lin, 2017). These features, as well as the regularization term, are the advantages of CSD over CE loss function (Janocha & Czarnecki, 2017; Kampa, Hasanbelliu, & Principe, 2011).

The presented analyses in addition to the further experimental results showed that the loss functions such as CSD and hinge are the reasonable alternatives for CE loss in OSV. However, the selection of the best loss completely relies on the application, dataset, and empirical results. The readers could find our empirical answer to the question of the best loss function in the further experiments section.

## 4. Integration of different losses into a unified multi-loss function

This section presents our ensemble approach to the central question of this study: How integration of different losses into a multi-loss function lead to an improved learning framework? We first investigate the characteristics of Snapshot Ensemble and multi-loss function. Due to their complementary features, we propose combining the features of both for addressing their limitations and overcome them by combining elements of the other method. Our proposed Multi-Loss Snapshot Ensemble (MLSE) is trained on genuine signatures based on the signature identification task. The representations learned by this framework is then employed for OSV.

### 4.1. Multi-loss function and Snapshot Ensemble, different but complementary approaches

- Snapshot Ensemble

In Snapshot Ensemble(Huang et al., 2017), as introduced earlier, each new trial initializes with trained weights of the previous trial. This relation between the sequential trials makes the optimization of new network faster because it eliminates the need for learning from scratch. This advantage significantly reduces the training cost of producing the ensemble. However, this method does not guarantee enough diversity between different snapshot networks while diversity is key for the success of ensemble(Brown et al., 2005; Masoudnia, Ebrahimpour, & Arani, 2012a). The authors claimed that this method could visit multiple good and diverse local minimums, leading to increasingly accurate predictions over the course of training in several classification benchmarks. However, the recently proposed study (Gotmare, Keskar, Xiong, & Socher, 2018) cast doubt on these claims according to the mode connectivity for loss landscape analysis. They found that the training process of Snapshot Ensemble is partly resilient to the changes in learning rates and argued against the claims about the converging to and escaping multiple local minima based on the theoretic analysis and experimental results. Furthermore, based on ensemble viewpoint, there were several findings (Brown et al., 2005; Kuncheva, 2014) against the technique of constructing an ensemble by changing learning rates or different initializations.

- Multi-loss function

Different loss functions lead the networks to reach different local minima. The analysis of different losses also showed they have their own strengths and limitations (Janocha & Czarnecki, 2017; Rosasco, De Vito, Caponnetto, Piana, & Verri, 2004; C. Xu et al., 2016), while the question arises as how to combine these advantages in a unified system without the heavy computational load required for independent runs of different loss functions and combining them. However, there is a concern about using a multi-loss function simultaneously in a DNN. It may reduce the training performance since different simultaneous supervisory signals could perturb the optimization process. In one point, simultaneous training of the network by different loss functions is not generally possible. In another point, training ensemble of networks independently using different losses brings a heavy computational burden.

## 4.2. Multi-Loss Snapshot Ensemble for feature ensemble learning

As it is clear from the analysis of the features of multi-loss function and Snapshot Ensemble and their advantages and disadvantages, these methods have different but complementary features. Based on the presented analysis, we propose a novel ensemble framework to include the advantages of both approaches as well as address their limitations. In the proposed Multi-Loss Snapshot Ensemble (MLSE), Snapshot Ensemble provides a framework to employ multi-loss function sequentially without the need for independent use of different loss functions in the training of different networks.

The MLSE framework includes a multi-trial training process in which a dominant loss function is selected for each trial. The dominant loss mainly contributes to the training of CNN while the other losses marginally contribute to the training due to their much smaller coefficients. The suggested Dynamic Multi-Loss (DML) function is proposed as:

$$\text{DML}(Y,O) = \lambda_1 \cdot \text{CE}(Y,O) + \lambda_2 \cdot \text{Hinge}(Y,O) + \lambda_3 \cdot \text{CSD}(Y,O) \tag{6}$$

where $\lambda_1$, $\lambda_2$, $\lambda_3$ are three numerical coefficients for these losses, respectively.

The derivative of the DML function with respect to the output is approximated as:

$$\frac{\partial_{\text{DML}(Y,O)}}{\partial_{o_j}} \simeq \lambda_1 \cdot \left(y_j - \text{pr}(o_j)\right) - \lambda_2 \cdot y_j \left[\max\left(0, \frac{1}{2} - y_j o_j\right)\right] + \lambda_3 \cdot \left[\left(y_j - \text{pr}(o_j)\right) - \|o_j\|\right] \tag{7}$$

where $\|o_j\|$ is the norm of $j$th output and acts as a regularization term.

The dynamic of DML function lies in the variation of $\lambda_i$ coefficients during training trials. In each trial, a loss function is selected as the dominant loss with a large $\lambda_i$ coefficient (0.98) and the other losses cooperate with a small $\lambda_j$ coefficient (0.02). The remaining loss functions act as the regularization terms for the dominant loss function. The dominant loss function in the sequential trials differ between CE loss, CSD loss and hinge loss, respectively. When training of the CNN network finishes on the first dominant loss, a snapshot is taken, and the weights of the trained network are saved as an ensemble member. The next loss function then becomes dominant, and training of the CNN continues on the second loss and so on. The proposed DML is employed for the training of a classic CNN with several convolutional layers followed by two Fully-Connected (FC) layers. All parameters of the CNN are shared between the proposed DML function except for the last FC layer while each loss function has a separate FC layer. A scheme of the proposed CNN structure is illustrated in Figure 3.

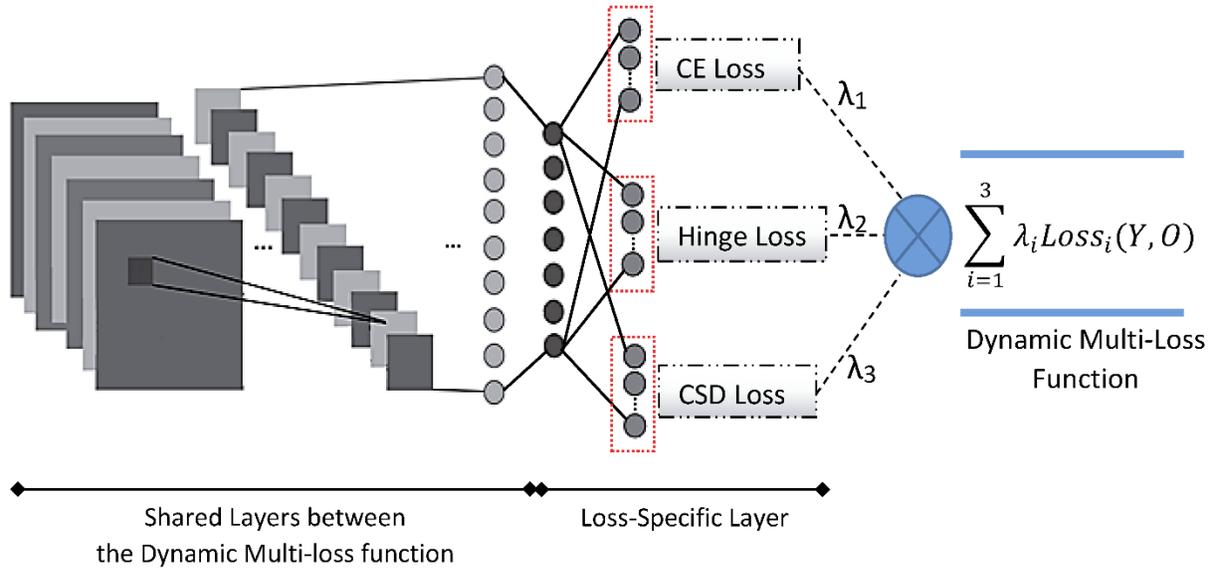

Fig. 3. Scheme of the proposed CNN structure for simultaneous using of different loss functions through the suggested dynamic multi-loss function. The basic losses used in the dynamic multi-loss function includes $loss_1$= CE, $loss_2$=Hinge and $loss_3$=CSD. A classic CNN with several convolutional layers followed by two FC layers is used. All parameters of the CNN are shared between the proposed multi-loss function except for the last FC layer while each loss function has a separate FC layer. The weights of each snapshot network are trained by minimizing the dynamic multi-loss function with different coefficients in the training trials.

We suggest the proposed MLSE framework for representation learning in the OSV problem. This framework first learns to classify genuine signature images based on the identification task. As a result, different snapshot CNNs are obtained, which then utilized for producing diverse multi-representation set from each input signature. A schematic diagram of this process is illustrated in Figure 4.a.

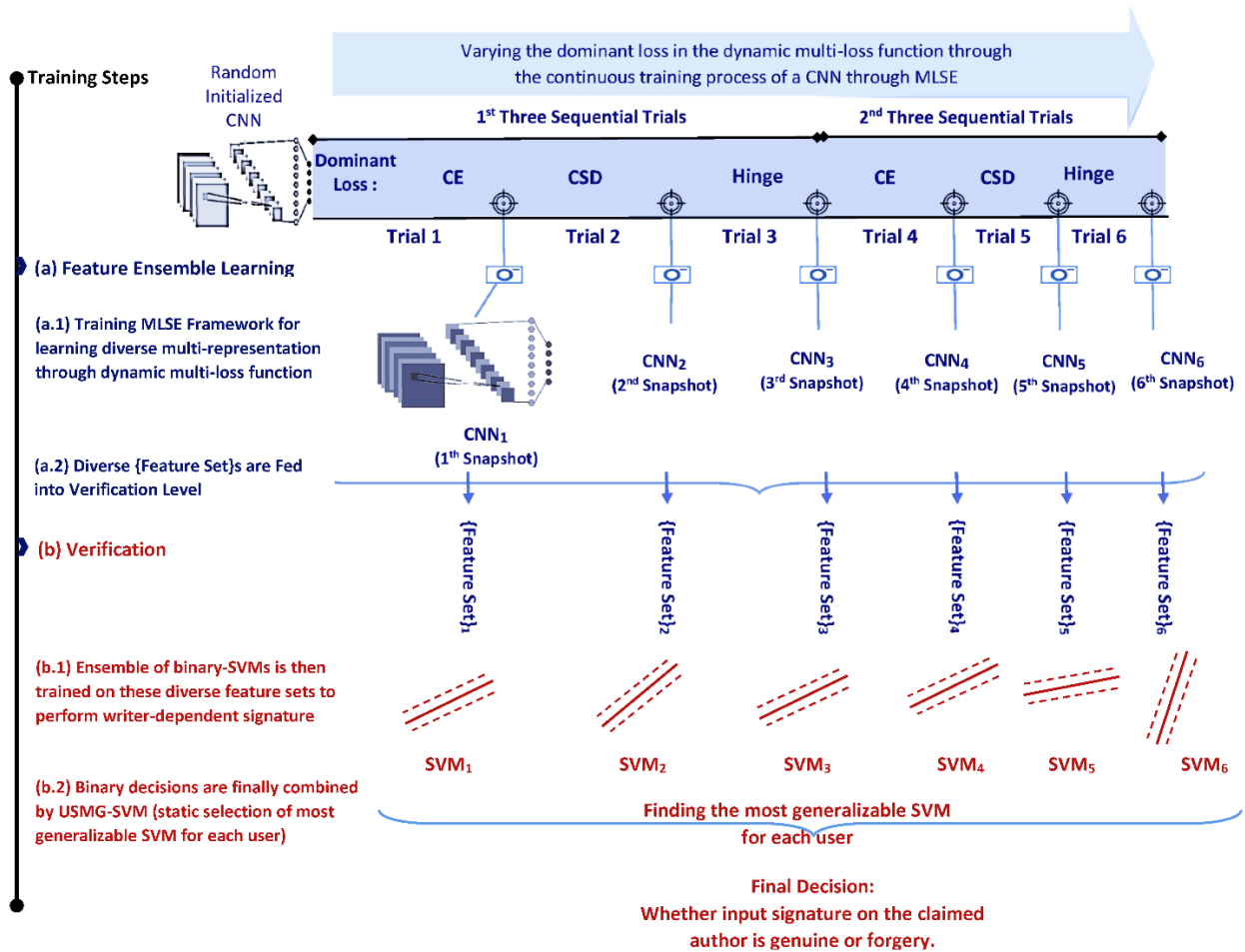

Fig 4. Sketch of our proposed OSV system. (a) Schematic diagram of the training process in the proposed MLSE framework for six trials. This framework produces six CNNs through the proposed DML function. In the target points, the trained CNN is saved, and the dominant loss function is then changed, and the training continues on the subsequent loss and so on. The target point at the end of each trial occurs when the networks' accuracy does not increase for five consecutive epochs. (b) Schematic diagram of the proposed verification method. Six diverse feature sets generated by MLSE are fed into the verification level. On each feature set, an SVM is trained and as a result, six SVMs are obtained for each user and their decisions are finally combined based on our proposed USMG-SVM method.

The number of sequential trials in the proposed MLSE framework can be set arbitrarily. However, as shown in Fig 4(a), we limit the number of sequential trials to 6 and as a result, six CNNs is produced. This number is empirically selected for the application of OSV. We will show in the experiments section that more trials and producing more than six CNNs do not improve the accuracy of OSV system significantly.

# 5. Verification by User-based Selection of Most Generalizable SVM

The proposed DML function in MLSE empowers the feature ensemble learning to learn diverse feature sets for each input signature. For each feature set, an SVM is then trained to classify the genuine signatures vs. random forgeries (the genuine signatures of other users). As a result, several SVMs are obtained for each user. Considering diversity between the feature sets and thus the learned SVMs, We suggest a selection approach to find a most appropriate SVM for each user among the ensemble of SVMs. The proposed static ensemble selection attempt to find the most generalizable SVM for each user. Due to a limited number of available signature images, we do not employ further validation set to estimate an expected generalization of each SVM in the ensemble. Instead, each trained SVM is evaluated on a different sampling of the genuine samples and different sets of random forgeries. To do so, data augmentation based on dropout (Konda, Bouthillier, Memisevic, & Vincent, 2015) is applied to the genuine samples and the set of random forgeries in each evaluation. These different conditions provide a testbed for estimating the expected generalization of each trained SVM and finding the best one for each user. We call this proposed selection method as User-based Selection of the Most Generalizable SVM (USMG-SVM). The pseudo-code 1 is presented that shows the details for USMG-SVM method.

---

**Pseudo-code 1**

Algorithm: Finding the best SVM in the ensemble for each user by USMG-SVM

Input: Six SVMs trained on the six diverse feature sets for each user in the training set.

Output: The index of best SVM for each user.

/* In all steps of the following algorithm, a 50%-dropout is applied on the feature sets before testing SVMs.

For each user {

    For five iterations{

    Random selection of several random forgeries (ten times the number of the genuine signatures).

    Applying a 50%-dropout on all feature sets of genuine signatures and random forgeries.

    For each of six SVMs in the ensemble{

    Testing the SVM on the two-class problem: genuine signatures for the user
    vs. random forgeries to estimate its expected generalization.

    }

    }

  }

  Index of the best SVM with the total maximum generalization for the user is achieved.

}

Return the Index of the best SVM for each user.

---

This method finds the index of best SVM for each user, and in the test phase, this index is used for the verification of input query signature based on the selected SVM among the ensemble. The schematic overview of the verification method is presented in Figure 4.b . Moreover, the flowcharts for the training and testing steps of the whole proposed OSV are illustrated in Figure 5.

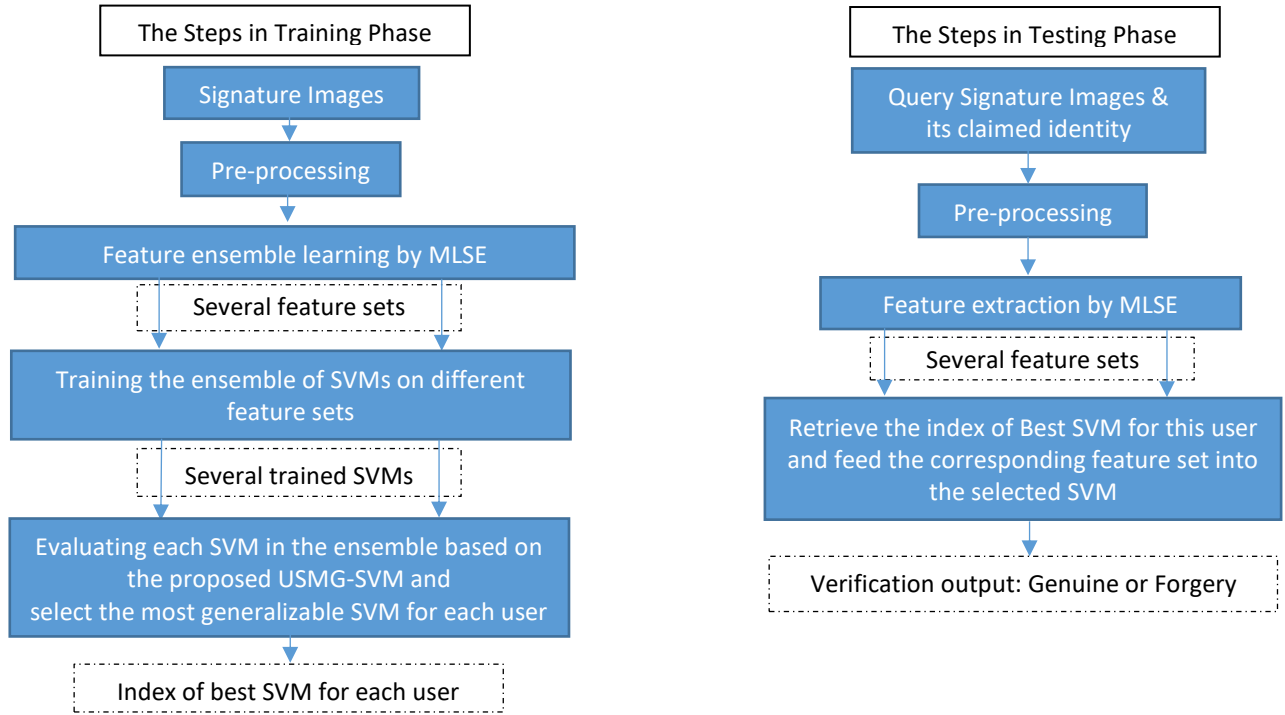

**Fig. 5.** Flowcharts for the training and testing steps of the proposed OSV.

# 6 Experimental Results

## 6.1 Datasets and Experimental Settings

Several experiments were conducted to analyze the performance of our proposed ensemble method for OSV and compare with the state-of-the-art results in the literature. We employed three offline signature datasets MCYT-75 (Fierrez-Aguilar, Alonso-Hermira, Moreno-Marquez, & Ortega-Garcia, 2004), UTSig (Soleimani et al., 2017) and GPDS-synthetic (Ferrer, Diaz-Cabrera, & Morales, 2015). Figure 6 shows some examples of signature images from these datasets. A summary of the datasets is given in Table 3.

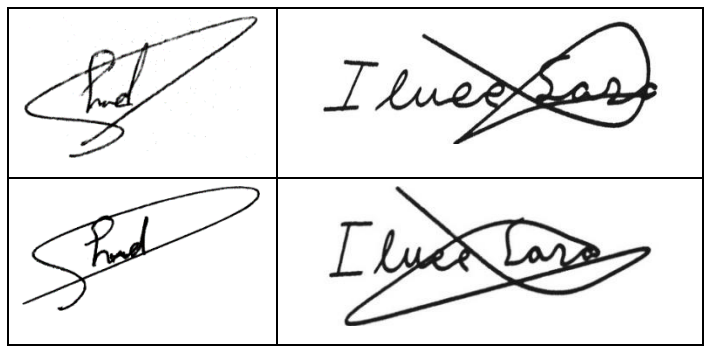

**Fig 6.** Some examples of signature images from UTSig and GPDS-synthetic. Each column shows two signatures of a user from these datasets, respectively. The images in the first row are the genuine signatures and the images in the second row are their corresponding skilled forgery signatures.

Table 3. Summary of the three signature datasets employed in this study.

| Dataset name | Nationality of offline signatures | #users and #images | #Genuine signatures | #Skilled forgeries |
|---|---|---|---|---|
| UT-Sig | Persian | 115 users with 8280 images | 27 | 42 |
| MCYT-75 | Spanish | 75 users with 2250 images | 15 | 15 |
| GPDS-synthetic | Synthesized signatures | 4000 users with 216000 images | 24 | 30 |

The signature images from the datasets need to be first pre-processed before the feature learning step. We followed the pre-processing approach suggested in (Hafemann et al., 2017a) which included removing the background by OTSU's algorithm (Otsu, 1979), inverting the image brightness, and resizing to the input size of the network.

The architecture used for the CNN in MLSE framework was partly inspired by (Hafemann et al., 2017a), which is described in Table 4. This architecture is composed of multiple layers of convolutions, max-pooling, and FC layers. The learnable layers including convolutional and FC layers are followed by Batch Normalization (Ioffe & Szegedy, 2015) and the Randomized leaky Rectified Linear Units (B. Xu, Wang, Chen,

& Li, 2015). The CNN network was initialized by He-Normal (He, Zhang, Ren, & Sun, 2015) and trained by Nesterov momentum (Botev, Lever, & Barber, 2017). The training hyper-parameters are listed in Table 5. As mentioned before, the trained ensemble of CNNs in the MLSE framework is used for feature learning in OSV. The outputs of last but one layer before the last FC layer of CNNs were considered as the feature sets for training the ensemble of SVMs in the verification step.

Table 4. The architecture of the CNN.

| Layer | Size | Other Parameters |
|---|---|---|
| Input + Dropout | Image size of the dataset | P= 0.1 dropout |
| Convolution | 96 x 11 x 11 | Stride=4, pad=0 |
| Max-Pooling | 96 x 3 x 3 | Stride=2 |
| Convolution | 256 x 5 x 5 | Stride=1, pad=2 |
| Max-Pooling | 256 x 3 x 3 | Stride=1, pad=2 |
| Convolution | 384 x 3 x 3 | Stride=1, pad=1 |
| Convolution | 384 x 3 x 3 | Stride=1, pad=1 |
| Convolution | 256 x 3 x 3 | Stride=1, pad=1 |
| Max-Pooling | 256 x 3 x 3 | Stride=2 |
| Fully Connected + Dropout | 2048 | $P_1$= 0.5 dropout |
| Fully Connected + Dropout | 2048 | $P_1$= 0.5 dropout |
| Independent Fully Connected Layers for each Loss Function + Softmax | 3 x C (C: Class Numbers) | 3 different loss functions (CE, CSD and hinge losses) |

Table 5. The training hyper-parameters in the MLSE framework for training the CNNs.

| Parameter | Value |
|---|---|
| Learning Rate (LR) | 0.01 |
| Momentum | 0.9 |
| Batch Size | 48 |
| λ coefficient for dominant loss | 0.98 |
| λ coefficient for regularizer losses | 0.02 |

Each SVM in the ensemble is trained on a two-class problem between genuine signatures of the user vs. genuine signatures of the other users as random forgeries. The binary-class balanced-SVM in Scikit-learn package was used to perform WD verification. Binary soft-margin linear SVMs in the one-vs-the-rest protocol with the default parameters (kernel = linear, C (Cost) = 1) were used in this work. The SVM balance factor was also ON in order for handling unbalanced classification. Based on this factor, the adaptation algorithm automatically adjusts weights inversely proportional between the sizes of two classes("Scikit-learn package,"). As explained before, the obtained ensemble of binary SVMs for each user was then combined based on the proposed USMG-SVM.

The mentioned protocol was used to set the parameters of the proposed algorithm for OSV. In all experiments (except for the parts 6.3 and 6.4) and for all signature datasets, 6 genuine images of each user were used for feature learning by the MLSE framework. In the verification step, the features extracted from these 6 images in addition to another 4 genuine images of the same user were used. Overall, a set including 10 genuine signatures for each user as positive samples and several sampled random forgeries as negative samples were used to train SVMs in the verification step. The size of random forgeries was determined ten times the number of genuine samples. Several experiments were conducted to analyze the performance of our proposed methods based on these experimental settings. Taking into account the differences in datasets and their protocols, we reported the number of signatures used for testing in Table 6.

Table 6. The training and testing sizes for the datasets.

| Dataset | Training Size (# genuine samples) | | Test Size (# samples) | |
|---|---|---|---|---|
| | Feature Learning | Verification (Ensemble of SVMs) | Genuine | Forgeries (Skilled) |
| UT-Sig | 6 | 4+(6) | 10 | 42 |
| MCYT-75 | 6 | 4+(6) | 5 | 15 |
| GPDS-synthetic | 6 | 4+(6) | 10 | 30 |

The performance of the proposed method was reported based on the False Rejection Rate (FRR), False Acceptance Rate (FAR), and Equal Error Rate (EER). FRR was calculated based on the rate of rejected genuine signatures and FAR according to the rates of accepted skilled forgeries. EER was also reported when FRR = FAR. All error rates were obtained by averaging ten different runs. We used repeated random sub-sampling for cross-validation. In each run, ten samples were randomly selected from the genuine signatures for training set while whole samples of the skilled forgeries were also used for the test set.

### 6.2 Empirical Answer to the question: which loss provides more generalization for feature learning in OSV?

We empirically approached the first asked question in our study. Several experiments were conducted to compare the performance of the independent runs of different loss functions in feature learning by the CNN. These experiments were performed on the three signature datasets: UT-Sig, MCYT, and GPDS-Synthetic with 6 training samples for each user. The results for the verification on these datasets are reported in Table 7.

The results showed that the FAR (RF) error rates for all loss functions were zero on both UT-Sig and MCYT datasets. For GPDS-Synthetic, the FAR (RF) was also almost zero. The minimum FAR error rate on random forgeries demonstrated that if these loss functions are used for training the CNNs in the identification task, the performances will be near perfect.

The other performance measures FAR, FRR and EER rates on the verification of skilled forgeries demonstrated different aspects of error in the OSV system. In most cases, using CE loss for feature learning exhibited the lowest error rates compared to CSD and Hinge losses. However, CSD loss also showed good performances so that in almost one-third of the cases, CSD loss reached the lower error rates compared to CE loss. Using Hinge loss for the feature learning in OSV also showed fairly good performances in most cases although it achieved the lowest error rate only in one case.

Table 7. The ERRs (%) of the OSV systems using different loss functions for feature learning by the CNNs in three signature datasets. For each dataset, independent training of the CNNs by different loss functions are compared based on different error measures: FAR on the Random Forgeries (RF) and FRR, FAR and EER on the Skilled Forgeries (SF). The least EER on each column are boldfaced. The reported EERs are shown for the average value (standard variation) of ten runs.

| Dataset / Loss Functions | FRR (SF) UT-Sig | FAR (RF) UT-Sig | FAR (SF) UT-Sig | EER (SF) UT-Sig | FRR (SF) MCYT | FAR (RF) MCYT | FAR (SF) MCYT | EER (SF) MCYT | FRR (SF) GPDS-S | FAR (RF) GPDS-S | FAR (SF) GPDS-S | EER (SF) GPDS-S |
|---|---|---|---|---|---|---|---|---|---|---|---|---|
| Hinge loss | **15.91 (0.52)** | 0 | 4.06 (1.19) | 8.06 (0.69) | 11.31 (0.77) | 0 | 1.68 (1.32) | 5.90 (0.86) | 17.12 (0.62) | 0.25 (0.09) | 10.99 (1.02) | 15.78 (0.77) |
| CE Loss | 17.61 (0.45) | 0 | 4.97 (0.78) | 7.63 (0.50) | **9.70 (0.49)** | 0 | 2.26 (0.94) | **5.46 (0.63)** | **10.62 (0.35)** | **0.11 (0.04)** | 7.73 (0.83) | 8.69 (0.36) |
| CSD Loss | 17.93 (0.31) | 0 | **3.14 (0.54)** | **7.45 (0.43)** | 14.92 (0.39) | 0 | **1.74 (0.65)** | 5.81 (0.56) | 12.41 (0.23) | 0.18 (0.03) | **8.20 (0.68)** | 10.90 (0.30) |

Therefore, there is not a unique solution in response to the mentioned question on the best loss function for feature learning in OSV. In other words, we could not find a loss function which performed the best for all signature datasets. However, the results showed the different competence of each loss function under different conditions. Thus, the significances of different loss functions in different datasets confirmed their complementary advantages for OSV problem.

## 6.3 Effect of Ensemble Size on the efficiency of the MLSE framework

The sensitivity of the MLSE framework to the variation of ensemble size is examined. In this experiment, we examined this framework on different ensemble sizes, including different numbers of sequential 3-trials. As a result, the ensemble sizes of three, six, nine and twelve were tested. The results are reported in Table 8.

**Table 8.** The ERRs (%) of our proposed OSV system with different numbers of sequential trials and as a result different ensemble sizes in three signature datasets. The MLSE framework was trained through the sequential trials, and the results of first, second, third and forth 3-trials, as well as the number of training epochs, are reported. At the end of each 3-trials, the results of combining by Majority Voting (MV) and USMG-SVM are also reported. Moreover, the result of each loss and the number of training epochs are also reported. The lowest EERs in each 3-trials are boldfaced. The reported EERs are shown for the average value (standard variation) of ten runs.

| Number of sequential trials | First 3-Trials | | Second 3-Trials | | Third 3-Trials | | Fourth 3-Trials | |
|---|---|---|---|---|---|---|---|---|
| Ensemble size | 3 (1*3) | | 6 (2*3) | | 9 (3*3) | | 12 (4*3) | |
| | EER(%) | #Epochs | EER(%) | #Epochs | EER(%) | Epochs | EER(%) | #Epochs |
| CE | 7.73 | 42 | 7.40 | 14 | 7.36 | 9 | 7.26 | 8 |
| Hinge | 8.51 | 13 | 7.93 | 11 | 7.89 | 11 | 7.90 | 10 |
| CSD | 7.82 | 15 | 7.32 | 12 | 7.30 | 13 | 7.32 | 11 |
| MV | 6.93 (0.73) | 70 (42+13+15) | 6.57 (0.31) | 37 (14+11+12) | 6.55 (0.28) | 33 (9+11+13) | 6.52 (0.30) | 29 (8+10+11) |
| USMG-SVM | 6.68 (0.61) | 70 | 6.17 (0.28) | 37 | 6.14 (0.24) | 33 | 6.15 (0.22) | 29 |

As the training goes on through the sequential 3-trials, the average number of epochs needed for convergence of each loss function was reduced. The training time required for each 3-trials also decreases as the training proceeds. It confirms the efficiency of training different losses one after another because the sequential training decreased the convergence time required for the subsequent losses. For example, the convergence time of hinge loss in the first 3-trials takes significantly fewer epochs compared to the previous CE loss. Moreover, the CE loss in the second 3-trials takes only 14 epochs for the convergence while the CE loss in the first one took significantly more times when initialized as the first loss.

As observed in the results, after the second 3-trials, we did not observe any significant improvements in the results. The similar results were also observed for MCYT and GPDS-Synthetic datasets. Therefore, we suggest using only two runs of 3-trials and thus six CNNs for our MLSE framework. Our proposed method with two sequential 3-trials takes 107 (70+37) epochs on average which is about 2.5 times of single loss (CE) training time (42 epochs). It confirms the time efficiency of our proposed MLSE because we produce six different CNNs in less than half of the time required for six independent CNNs.

## 6.4. Efficiency and robustness of the MLSE framework in different training sizes

The sensitivity of the proposed feature ensemble learning to the variation of the training size is empirically tested. In this experiment, the number of training signature images used for feature learning by MLSE framework changes, N= 4, 6, 8, and 10. The number of <u>4 + N</u> training samples were also used for training the ensemble of SVMs in the verification level. Moreover, the performance of MLSE framework combined by the USMG-SVM was compared to the best loss function (from part 6.2) and a traditional combining technique, i.e., majority voting. The results are reported in Table 9.

Predictably, the results showed that with increasing the numbers of training samples used for feature learning, the error rates gradually decreased. The low error rates on the different training sizes also confirmed the robustness of the MLSE framework against the small training size. The significant performance improvements with the proposed feature ensemble learning (whether combined by majority voting or USMG-SVM) over the best single loss were also observed. The superiority of MLSE framework over the single loss empirically confirmed our hypothesis about the complementary advantages of different loss functions and the efficiency of our proposed dynamic multi-loss function. In other words, the low error rates, even with small training size (N=4), approved the efficiency of our proposed diversity-induced regularization in the generalization of MLSE framework.

**Table 9.** The ERRs (%) of the OSV systems when different training sizes used for feature learning in the MLSE framework are reported for three signature datasets. The results also include the Single loss error rate in addition to the Majority Voting (MV) and USMG-SVM combining methods. The lowest EERs on each row are boldfaced. The reported EERs are shown for the average value (standard variation) of ten runs.

| Dataset / Training Size | EER (%) | | | | | | | | |
|---|---|---|---|---|---|---|---|---|---|
| | Best Single loss | MLSE combined by MV | MLSE combined by USMG-SVM | Best Single loss | MLSE combined by MV | MLSE combined by USMG-SVM | Best Single loss | MLSE combined by MV | MLSE combined by USMG-SVM |
| | UT-Sig | UT-Sig | UT-Sig | MCYT | MCYT | MCYT | GPDS-S | GPDS-S | GPDS-S |
| 4 | 9.78 (0.95) | 8.24 (0.47) | **7.91 (0.35)** | 6.90 (1.30) | 5.09 (0.94) | **4.37 (0.68)** | 9.77 (0.98) | 7.62 (0.69) | **6.70 (0.58)** |
| 6 | 7.45 (0.43) | 6.57 (0.31) | **6.17 (0.28)** | 5.46 (0.63) | 4.10 (0.47) | **2.93 (0.40)** | 8.69 (0.37) | 6.47 (0.30) | **6.13 (0.25)** |
| 8 | 7.24 (0.38) | 6.20 (0.22) | **5.48 (0.22)** | 4.90 (0.50) | 3.63 (0.39) | **2.41 (0.33)** | 7.93 (0.31) | 6.31 (0.26) | **5.59 (0.20)** |
| 10 | 7.09 (0.33) | 5.95 (0.16) | **5.32 (0.10)** | 4.74 (0.44) | 3.17 (0.26) | **1.99 (0.18)** | 7.55 (0.29) | 6.05 (0.21) | **5.48 (0.14)** |

Moreover, we compared the performance gain achieved by our proposed combiner method USMG-SVM with majority voting. This comparison showed that USMG-SVM achieved fewer error rates for different training sizes and all datasets, however, its superiority was not significant in few cases.

### 6.5 Efficiency and robustness in the arrival of new users enrolled in the verification system

We conducted this experiment to examine the efficiency and robustness in the arrival of new users enrolled in the verification system. Based on this protocol, for each dataset, 20% of the users were randomly selected for feature learning while the remaining users used for evaluating the generalization of the verification system. Next, these two sets are replaced and the second set is used for feature learning and the first one for the performance evaluation. In fact, the suggested protocol leads our proposed method to WI feature learning while the previous protocol (considering all users) used WD feature learning. However, the verification in both protocols is performed based on WD verification.

This random cross-validation (in term of users) is repeated 5 times independently, and the final results were obtained based on averaging 10 trials. In this experiment, 10 genuine signatures of each user in the training set are used for feature learning by MLSE while 10 genuine signatures of the users in addition to all skilled forgeries in the test set are also used for verification. The results are shown in Table 9 and compared to the previous protocol in which all enrolled users are available for training.

Table 10. The ERRs (%) of the OSV systems in the arrival of new users enrolled in the verification system. The results are compared to the previous protocol in which all enrolled users are available for training. The results are reported for three signature datasets. The results also include the Single loss error rate in addition to the Majority Voting (MV) and USMG-SVM combining methods. The lowest EERs on each row are boldfaced. The reported EERs are shown for the average value (standard variation) of ten runs.

|  | EER (%) | | | | | | | | |
|---|---|---|---|---|---|---|---|---|---|
|  | Best Single loss | MLSE combined by MV | MLSE combined by USMG-SVM | Best Single loss | MLSE combined by MV | MLSE combined by USMG-SVM | Best Single loss | MLSE combined by MV | MLSE combined by USMG-SVM |
|  | UT-Sig | UT-Sig | UT-Sig | MCYT | MCYT | MCYT | GPDS-S | GPDS-S | GPDS-S |
| WD feature learning | 7.09 (0.33) | 5.95 (0.16) | **5.32 (0.10)** | 4.74 (0.44) | 3.17 (0.26) | **1.99 (0.18)** | 7.55 (0.29) | 6.05 (0.21) | **5.48 (0.14)** |
| WI feature learning | 10.39 (1.64) | 7.66 (0.52) | **7.02 (0.39)** | 8.69 (1.90) | 6.53 (0.87) | **5.85 (0.71)** | 10.28 (2.02) | 7.41 (1.13) | **6.57 (0.83)** |

In MCYT, with the least number of users and images compared to the other datasets, the obtained error 5.85% was significantly larger than our previous result 1.99%. In UTSIG, with a few number of users and images, the obtained error 7.02% was not significantly larger the previous result 5.32%. In GPDS-Synthetic, with a huge number of users and images, the obtained error 6.57% was comparable to the previous result 5.48%. As the size of the datasets increased, we observed a lower increase in the error rates in the comparison of two protocols. These effects were observed since the presence of all users to the CNNs leads them to better feature learning. In the larger datasets, we observed less drop in the performances because the availability of more users helps the CNNs to learn more diverse patterns and partially compensates the lack of other users.

The overall results of our proposed OSV system in the protocol of WI feature learning are comparable with the protocol of WD feature learning in the two larger datasets. It confirms the generalization ability of our proposed method in the arrival of new users enrolled in the OSV system.

### 6.6 Comparison with the state-of-the-art in different signature datasets

The state-of-the-art performances on the three signature datasets are compared with our proposed method in this part. Table 11-13 present the comparison with the state-of-the-art performance on UT-Sig, GPDS-Synthetic, and MCYT, respectively.

On UT-Sig, we achieved EER of 5.32% which is substantially better than the best reported EER of 17.45% by (Soleimani et al., 2016) in the literature. Since this dataset has recently published (Soleimani et al., 2017), few studies (Soleimani et al., 2016; Soleimani et al., 2017) applied their methods on it. Therefore, we applied state-of-the-art methods, including SigNet, and SigNet-F (Hafemann et al., 2017a) and also Snapshot Ensemble to learn features on UT-Sig dataset in order for fair comparison. As shown in Table 11, Our proposed method could also achieve better performance compared to these methods by significant margins.

For GPDS-Synthetic, there was also a few studies (Ferrer et al., 2015; Serdouk et al., 2017; Soleimani et al., 2016) examining whole signatures of the 4000 users in the dataset. In this dataset, we achieved EER of 5.48% which is a substantial improvement over the best reported EER of 13.3% (Soleimani et al., 2016) in the literature. Moreover, we applied state-of-the-art methods, including SigNet, and SigNet-F (Hafemann et al., 2017a) and also Snapshot Ensemble to learn features on GPDS-Synthetic dataset in order for fair comparison. The results in Table 12 confirms that our proposed MLSE framework combined with USMG-SVM obtains significantly better performance compared to the tested methods.

For MCYT, there were very studies on this data in which the recent study (Hafemann et al., 2017a) reported very good accuracies, as shown in Table 13. We achieved EER of 2.93% which is comparable to the best EER of 2.87% in the literature. However, this state-of-the-art EER was achieved using GPDS-960 dataset which is no longer available according to the announcement of its provider ("GPDS Signature database,"). Putting it all together, the results confirm that our proposed feature ensemble learning and the ensemble of SVMs combined through USMG-SVM approach could achieve the state-of-the-art results on UT-Sig and GPDS-Synthetic datasets and also comparable performance on MCYT dataset.

Moreover, our results in WI feature learning protocol are comparable to WD feature learning protocol. It confirms the generalization ability of our proposed method in the arrival of new users enrolled to the OSV system.

The overall results we achieved as well as the results of other CNN-based approaches also confirms the superiority of feature learning approaches compared to hand-crafted features. It shows that CNNs in general and the ensemble of CNNs in specific can efficiently handle the inherent challenges of OSV, including distortion in signatures, high intra-personal variation, and low inter-personal variation.

Table 11. Comparison with the state-of-the-art in UTSig.

| Reference | Feature | | Classifier | | #Samples per User | EER (%) |
|---|---|---|---|---|---|---|
| | Method | Type | Method | Type | | |
| (Soleimani et al., 2016) | DRT+DMML | WD | Thresholding | WD | 12 | 20.28 |
| (Soleimani et al., 2016) | HOG+DMML | WD | Thresholding | WD | 12 | 17.45 |
| (Soleimani et al., 2017) | Fixed-point geometrics | WI | SVM | WD | 12 | 29.71 (0.29) |
| (Younesian et al., 2019) | ResNet CNN pretrained on ImageNet | WI | Active learning with SVM | WD | 7 | 16.40 |
| (Mersa et al., 2019) | ResNet CNN pretrained on Handwriting classification tasks | WI | SVM | WD | 5 / 7 / 10 | 11.16 / 9.96 / 9.80 |
| (Hafemann et al., 2017a) (our test[1]) | SigNet | WI | SVM | WD | 10 | 9.61 (0.21) |
| (Hafemann et al., 2017a) (our test[1]) | SigNet-F | WI | SVM | WD | 10 | 10.54 (0.29) |
| (Huang et al., 2017) (our test[1]) | Snapshot Ensemble of CNNs | WD | Ensemble of SVMs + USMG-SVM | WD | 10 | 9.80 (0.60) |
| **Present Work** | MLSE | WI / WD | Ensemble of SVMs + USMG-SVM | WD | 10 | 7.02(0.39) / 6.17(0.28) |

---

[1] We applied this method on UT-Sig dataset and reported the result.

Table 12. Comparison with the state-of-the-art in GPDS-Synthetic.

| Reference | Feature | | Classifier | | #Overall Used Users | #Samples per User | EER (%) |
|---|---|---|---|---|---|---|---|
| | Method | Type | Method | Type | | | |
| (Soleimani et al., 2016) | HOG + DMML | WI | Thresholding | WD | 2500 | 10 | 12.80 |
| | | | | | 4000 | | 13.30 |
| (Ferrer et al., 2015) | LBP | WI | SVM | WI | 4000 | 10 | 16.44 |
| (Serdouk et al., 2017) | HOT | WI | AIRSV | WD | 4000 | 10 | 16.68 |
| (Zhang et al., 2016) | GAN | WI | Thresholding | WD | 4000 | 10 | 14.79 |
| (Huang et al., 2017) (our test[2]) | Snapshot Ensemble of CNNs | WD | Ensemble of SVMs + USMG-SVM | WD | 4000 | 10 | 11.02 (0.61) |
| (Hafemann et al., 2017a) (our test[2]) | SigNet | WI | SVM | WD | 4000 | 10 | 9.53 (0.40) |
| (Hafemann et al., 2017a) (our test[2]) | SigNet-F | WI | SVM | WD | 4000 | 10 | 8.70 (0.35) |
| Present Work | MLSE | WI / WD | Ensemble of SVMs + USMG-SVM | WD | 4000 | 10 | 6.57(0.83) / 6.13(0.25) |

---

[2] We applied this method on GPDS-Synthetic dataset and reported the result.

Table 13. Comparison with the state-of-the-art in MCYT.

| Reference | Feature | | Classifier | | #Samples per User | EER (%) |
|---|---|---|---|---|---|---|
| | Method | Type | Method | Type | | |
| (Vargas et al., 2011) | LBP | WI | SVM | WD | 10 | 7.08 |
| (Ooi et al., 2016) | DRT+PCA | WI | PNN | WD | 10 | 9.87 |
| (Soleimani et al., 2016) | HOG + DMML | WD | Thresholding | WD | 10 | 9.86 |
| (E. N. Zois et al., 2016)[3] | Poset-oriented grid features | WI | SVM | WD | 5 | 6.02 |
| | | | | | 10 | 4.01 |
| (Hafemann et al., 2017a) | SigNet | WI | SVM | WD | 5 | 3.58(0.54) |
| | | | | | 10 | 2.87(0.42) |
| (Hafemann et al., 2017a) | SigNet-F | WI | SVM | WD | 5 | 3.70(0.79) |
| | | | | | 10 | 3(0.56) |
| (Hafemann et al., 2018) | SigNet-SPP | WI | SVM | WD | 10 | 3.4(1.08) |
| (Sharif et al., 2018) | Global + local descriptors + feature selection by GA | WD | SVM | WD | 5 | 9.16 |
| | | | | | 10 | 7.92 |
| | | | | | 16 | 3.75 |
| (E. N. Zois et al., 2018)[4] | Dictionary learning | WI | Thresholding | WD | 5 | 3.52 |
| (Okawa, 2018a) | BoVW and VLAD | WI | SVM | WD | 10 | 6.4 |
| (Okawa, 2018b) | Fisher vector | WI | SVM | WD | 10 | 5.47(1.18) |
| (Mersa et al., 2019) | ResNet CNN pretrained on Handwriting classification tasks | WI | SVM | WD | 5 | 7.12 |
| | | | | | 7 | 5.51 |
| | | | | | 10 | 3.98 |
| (Huang et al., 2017) (our test[5]) | Snapshot Ensemble of CNNs | WD | Ensemble of SVMs + USMG-SVM | WD | 10 | 5.80(1.3) |
| Present Work | MLSE | WI | Ensemble of SVMs + USMG-SVM | WD | 10 | 5.85(0.71) |
| | | WD | | | | 2.93(0.40) |

---

[3],[4] Unlike the most studies in the table that used only skilled forgeries for verification, this study used combination of random and skilled forgeries.

[5] We applied this method on MCYT dataset and reported the result.

## 7. Conclusion and Future Works

Our novel approach to OSV originated with studying two asked questions about the best loss and the ensemble of losses for feature learning in OSV. Between the loss functions examined in this research including CE, CSD and Hinge loss, no one could achieve the best accuracy in all datasets. Instead, we suggested using the dynamic multi-loss function in Snapshot Ensemble to simultaneously utilize all loss functions in the continuous training of MLSE framework. This framework could combine the different but complementary advantages of different loss functions and produced rich, diverse feature sets. In this way, the proposed framework supports the need of OSV for the diversity-induced regularization to tackle the challenges of this task. In the verification step, the ensemble of SVMs and a novel static selection approach were suggested. The experiments with UT-Sig, GPDS-Synthetic, and MCYT demonstrated that the proposed verification system could surpass the state-of-the-art performance on the first two datasets and comparable performance in the last one. Hence, the experimental results in addition to the theoretical discussion on the challenges of OSV proved that our proposed method could handle these challenges by exploitation of the suggested diversity and regularization in the ensemble.

For future work, we are going to examine other diversity-induced regularization factors in MLSE framework and also extend its applications. Using the proposed framework for more challenging tasks such as cross-database OSV is one of the application that may challenge the generalization ability of ours. Regardless of the OSV task, the proposed MLSE framework provides a general approach for using the simultaneous multi-loss function in different verification applications as well as classification tasks. It will be further studied in our future work.

## Acknowledgment

This work was supported by the Ph.D. grant from Cognitive Sciences and Technologies Council of IRAN and the Ph.D. grant form NBIC research center of University of Tehran.